\documentclass{article}
\usepackage[utf8]{inputenc}

\title{Artistic Domain Generalisation Methods are Limited by their Deep Representations}
\author{Padraig Boulton and Peter Hall}

\date{April 2019}

\usepackage[a4paper, top=20mm, bottom=20mm, left=20mm, right=20mm]{geometry}
\usepackage{arydshln}
\usepackage{amsmath}
\usepackage{algorithm}
\usepackage[noend]{algpseudocode}
\usepackage{placeins}
\usepackage{subfigure}
\usepackage{comment}
\usepackage{graphicx}

\DeclareMathOperator*{\argmax}{arg\,max}

\def\etal{\emph{et al.}}

\begin{document}

\maketitle

\begin{abstract}
The cross-depiction problem refers to the task of recognising visual objects regardless of their depictions; whether photographed, painted, sketched, {\em etc}. In the past, some researchers considered cross-depiction to be domain adaptation (DA). More recent work considers cross-depiction as domain generalisation (DG), in which algorithms extend recognition from one set of domains (such as photographs and coloured artwork) to another (such as sketches). We show that fixing the last layer of AlexNet to random values provides a performance comparable to  state of the art DA and DG algorithms, when tested over the PACS benchmark. With support from background literature,  our results lead us to conclude that texture alone is insufficient to support generalisation; rather, higher-order representations such as structure and shape are necessary.


\end{abstract}

\section{Introduction}
\label{sec:intro}

Humans can recognise objects across an incredibly wide variety of  different depiction styles; a dog is a dog is a dog whether photographed by a professional, painted by an amateur, or drawn by a child. Unfortunately,  the same does not appear to be true for most recognition algorithms: most exhibit a fall in performance when presented with depictions of objects in even a moderately diverse set of depiction styles \cite{Hall2015,dbadg}.  We call this the {\em cross-depiction problem}.

The cross-depiction problem is under-researched, yet is significant because it clearly demonstrates the limitations of current approaches. As a matter of principle, it brings into question assumptions, tacit or otherwise, that are common to many algorithms. As a matter of practice, a machine able to recognise with the versatility of humans would undoubtedly enhance applications such as image-based search \cite{artofdetection, Crowley14a}, sketch-based image retrieval \cite{collomosse_sketch, Collomosse_2017_ICCV}, image-to-image translation~\cite{Gatys2016}, and possibly support research in areas such as human cognition.

The difficulty of the cross-depiction problem has two roots. First, the fact that object identity remains constant, for humans, under a much wider variability than is normally the case for the datasets used in Computer Vision~\cite{2015beyond,dbadg}. Second, the fact humans can generalise  to previously unseen styles ({\em e.g.} people understand figurative art outside their mother culture, see castles in clouds, and so on). This has led many researchers to argue that the cross-depiction problem is one of domain adaptation~\cite{Hall2015}, or more recently of domain-generalisation~\cite{dbadg,metareg,mldg}.

In this paper, we provide evidence that domain-generalisation as currently conceived in state-of-the-art literature is not sufficient to address the cross-depiction problem. In particular, we show that an AlexNet classifier with fixed random weights in the final  fully-connected layer compares well to the performances of two contemporary meta-learning algorithms (MetaReg~\cite{metareg} and MLDG~\cite{mldg}), both of which have been used to address cross-depiction. 

Details of our network, along with an intuitive explanation, can be found in Section~\ref{sec:method}, followed by experimental results in Section~\ref{sec:expts}. These results, with evidence from the literature is used to argue that there is an over-reliance on texture in contemporary recognition approaches~\cite{texturebias,Wu2014} and that this extends to learning measures over a texture space, see Section~\ref{sec:conc}. The implication is that to succeed in the cross-depiction problem, research has to address problems of representation, not only of learning measure.

\section{Related Work}
\label{sec:relwork}


The cross-depiction problem refers to the task of recognising visual objects regardless of their depiction whether realistic or artistic. It is an under-researched area. Some work uses constellation models, {\em e.g.} Crowley and Zisserman use a DPM to learn figurative art on Greek vases~\cite{crowley2013gods}. Others develop the problem of searching a database of photographs based on a sketch query; edge-based HOG was explored in \cite{hu-cviu13}. Li {\emph et al.}~\cite{li2013sketch} developed rich sketch representations for sketch matching using both local features and global structures of sketches. Others have investigated sketch based retrieval of video ~\cite{hu2013markov,collomosse2009storyboard}. Wu {\em et al} ~\cite{Wu2014} provide a non-neural fully-connected constellation model that is stable across depictions.

Deep learning has recently emerged as a truly significant development in Computer Vision. It has been successful on conventional databases, and over a wide range of tasks, with recognition rates in excess of $90$\%. Deep learning has been used for the cross-depiction problem, but its success is less clear cut. Crowley and Zisserman~\cite{Crowley14a,artofdetection} are able to retrieve paintings in 10 classes at a success rate that does not rise above 55\%; their classes do not include people. Ginosar {\em et al}~\cite{Ginosar_etal_eccvvisart2014} use deep learning for detecting people in Picasso paintings, achieving rates of about 10\%.

Other than this paper, we know of only two studies aimed at assessing the performance of well established methods on the cross depiction problem.  Crowley and Zisserman~\cite{CrowleyZisserman_bmvc2014} use a subset of the `Your Paintings' dataset~\cite{yourpaintings}, the subset decided by those that have been tagged with VOC categories~\cite{everingham2010pascal}. Using 11 classes, and objects that can only scale and translate, they report an overall drop in per class Prec@k (at $k=5$) from 0.98 when trained and tested on paintings alone, to 0.66 when trained on photographs and tested on paintings. Hu and Collomosse~\cite{hu-cviu13} use 33 shape categories in Flickr to compare a range of descriptors: SIFT, multi-resolution HOG, Self Similarity, Shape Context, Structure Tensor, and (their contribution) Gradient Field HOG. They  test a collection of 8 distance measures, reporting low mean average precision rates in all cases. Our focus is on domain-shift via meta-learning, and therefore concentrate our review on that area.


Datasets exhibit bias, which can be problematic. In photographic image recognition, bias for particular camera settings and other attributes can prevent models generalising well \cite{bias}. This motivated the collection of the multi-domain VLCS dataset: an aggregation of photos from Caltech, LabelMe, Pascal VOC 2007 and SUN09 \cite{bias}. Until recently, domain adaptation and generalisation in image recognition focused on transfer across photo-only benchmarks. Now, more datasets are available that cover larger domains shift across more varying depictive styles \cite{Wu2014,dbadg,Wilber_2017_ICCV} and better reflect the cross-depiction problem. We make use of the PACS dataset provided by Li \etal \cite{dbadg}. As a domain generalisation benchmark, where one domain is an unseen target domain, PACS is a far more challenging task than photographic benchmarks. Li \etal \cite{dbadg} measured an average KL-divergence \cite{kld} of 0.85 between training and test domains across PACS, compared to 0.07 across VLCS.

Domain Adaptation (DA) attempts to compensate for bias by adapting a model constructed on one domain to a target domain using examples from that new domain, {\em e.g.} \cite{Saenko:2010}. DA has been used in the cross depiction problem with both non-neural~\cite{2015beyond} and neural algorithms, such as the Domain Separation Network (DSN) \cite{dsn}.

Recently, Domain Generalisation (DG) approaches have gained attention. These differ from DA in that DG algorithms have no access to the target domain. General approaches include learning domain invariant representations, or  deriving domain agnostic classifiers by assuming individual domains' classifiers consist of domain-specific and domain-agnostic components, then extracting the latter \cite{undobias}. Examples of relevance here are   Domain Multi-Task Auto Encoders (D-MTAE) \cite{dmtae} and ``Deeper-Broader-Artier'' network (DBA-DG) \cite{dbadg}. Most recently, MetaReg~\cite{metareg} and MLDG~\cite{mldg} exhibit state of the art performance on the PACS dataset.

Meta-Learning and ``learning to learn'' have been a part of machine learning for a long time \cite{Schmidhuber95, Schmidhuber1997, Thrun:1998}, and remains relevant in deep learning. Learning to learn is strongly related to DA and DG. DG research of relevance here tends to treat domains as discrete; meta-learning approaches use multiple domain-specific classifiers \cite{metareg} or optimisation steps \cite{mldg}. Effectively, they learn domain generalising models by completing domain adaptation internally within the set of training domains. 



We show that contemporary DG classifiers that employ meta-learning perform no better than an AlexNet furnished with fixed random weights for its final fully-connected layer. Random projection is a universal sampling strategy that separates data according to the angles between points \cite{random, Qiu2013, Yamaguchi}. The weights of a network represent a projection which, when trained, adapt to preserve certain distances over others and prioritise minimizing intra-class angles over maximising inter-class ones \cite{random}. Thus in domain generalisation, a layer may learn latent domain-specific embeddings rather than a single, generalising projection. Keeping fixed random weights reduces internal domain-specific learning.



\section{Experiments}
\label{sec:method}

Our hypothesis is that object class representation is the key to the cross depiction problem. Meta-learning algorithms are intended to generalise by learning a measure between classes, rather than make use of a prescribed measure. Our experiment is designed to test the efficacy of meta-learning algorithms that have been used on the cross-depiction problem, when compared to a much simpler algorithm with a prescribed measure.  The advanced algorithms we experiment with are Meta-Learning for Domain Generalisation (MLDG) \cite{mldg} and Meta-Regularisation MetaReg \cite{metareg}. Our ``simple'' network is AlexNet ~\cite{alexnet} with the final layer weights fixed to random values. Pseudo-code for these networks can be found in Algorithms~\ref{alg:mldg}, \ref{alg:metareg}, and \ref{alg:us}, respectively. The database we use to test on is PACS (Photographs, Artwork, Cartoon, Sketch) \cite{dbadg} dataset.

\subsection{Dataset}
\label{subsec:PACS}

\begin{figure}[]
\includegraphics[height=3cm]{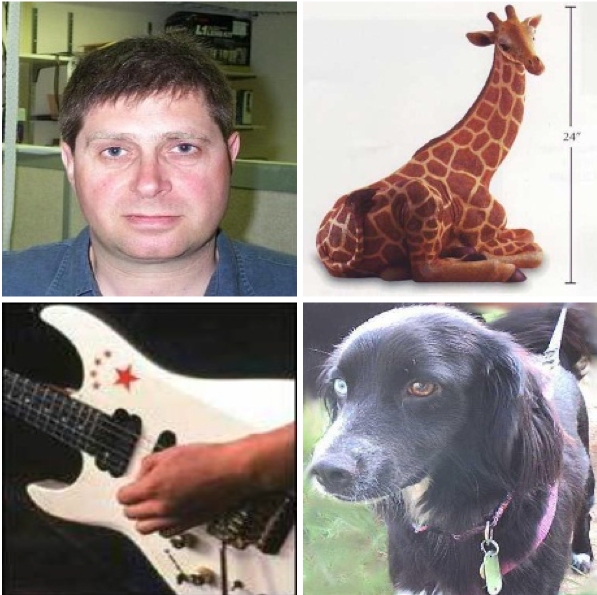}
\includegraphics[height=3cm]{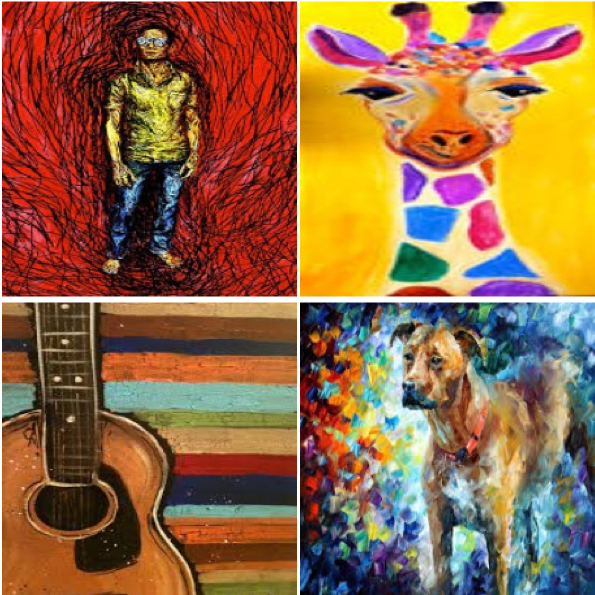}
\includegraphics[height=3cm]{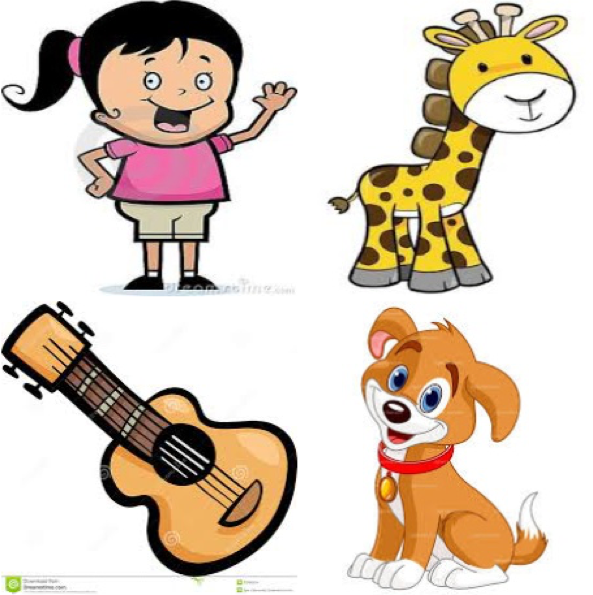}
\includegraphics[height=3cm]{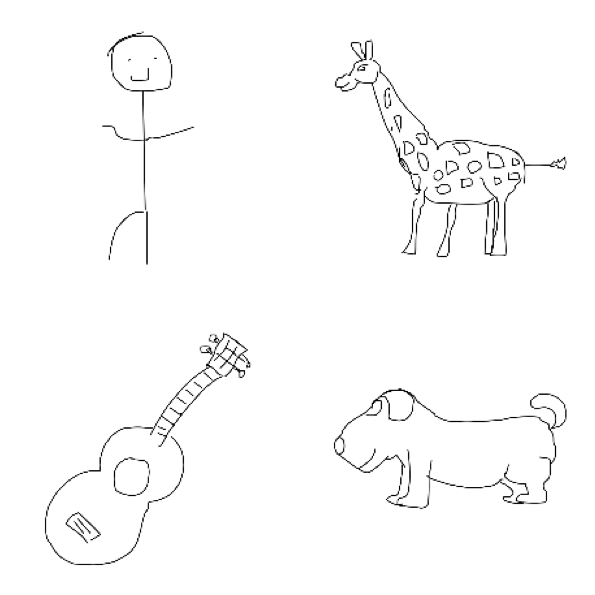}
\caption{Examples from the PACS dateset \protect{\cite{dbadg}}. Left to right: Photo, Art, Cartoon, Sketch.}
\label{fig:PACS}
\end{figure}

PACS contains four domains - Photo, Art painting, Cartoon and Sketch. Examples are shown in Figure~\ref{fig:PACS}. Performance over PACS is measured in four scenarios. In a given scenario, one domain is an unseen target domain for testing. The three remaining domains are source domains used for training the model. Li \etal \cite{dbadg} proposed PACS as a domain generalisation problem that reflects how the quantity of images available can be fundamentally constrained for some depictive styles \cite{dbadg}. Lack of examples may prevent training good models for certain styles; domain generalisation learns these by leveraging information from available styles. 

\subsection{Domain Generalisation Algorithms}

We wish to experiment with state of the art algorithms for domain generalisation applied to cross-depiction.  Li \etal \cite{dbadg} report that end-to-end trained models outperform those with pre-trained deep features.   MLDG~\cite{mldg} and MetaReg~\cite{metareg} are both trained end-to-end, and both achieve state of the art on PACS through meta-learning. We now describe each of them in a little more detail.

These meta-learning algorithms partition PACS into three parts, each comprising one or more of its domains. The ``meta-train'' set is used to train a network in the conventional way, a ``meta-test'' set is akin to a validation set that is used to update meta-parameters, while a hold-out set is used to simulate an unseen domain. For example, the P and A domains may be the meta-train set, the C domain the meta-test set, and P the hold-out set. The meta-train and meta-test set are the only domains used during learning, the hold-out set is used to test the full trained network and plays no role at all in learning.


\paragraph{MLDG} learns optimisation updates that minimise losses over the  meta-train and meta-test domains in a coordinated way. Broadly, network parameters $\Theta$ are updated using two loss functions: $F(.)$ over the meta-train set and $G(.)$ over the meta-test set. The gradient of $F(\Theta)$ is used to estimate a new parameter vector $\Theta'$; this new parameter is then used to update the original using the combined gradient $\Theta \leftarrow \Theta - \gamma \partial[ F(\Theta) + G(\Theta')]$ (note the use of two different model parameters). See Algorithm~\ref{alg:mldg} for further details.

\paragraph{MetaReg} harmonises a collection of classifiers, each of which are fed from a common convolutional network and each of which feed a common regularising network. Without the regularisor, each network would be trained as a conventional neural classifier, and such a step exists as the first stage of MetaReg. The second step then randomly selects a pair of domains, one of which is updated using the regularisor (meta-train), the second of which is used to update the regularisor (meta-test). Each iteration selects a new random pair of domains until the regularisor is fully trained. As this point the regularisor is fixed and the final task network is trained {\em ab initio} on the aggregation of source domains. Algorithm~\ref{alg:metareg} provides psuedo-code for the meta-learning stages of MetaReg.

\subsection{A Fixed Random Weight Network}\label{approach}

Our network has been motivated from our observations on the work of Collomosse \etal ~\cite{collomosse_sketch,Collomosse_2017_ICCV}. Those authors were interested in sketch based visual search; they used contrastive loss under a triplet network to disentangle content from style. In their context, contrastive loss attempts to project examples of a single class onto a single point in space regardless of style {\em i.e.} a domain-invariant representation of object class/content but not style. Applying contrastive loss to artistic depictions took considerable effort, and other previous research suggests it is better to allow variation due to depiction \cite{Wu2014}. With this in mind, we choose dominant direction as class indicator rather than location in space to allow variation due to depiction. Provided the dominant direction is correct, the relative magnitude of a vectors' components are irrelevant.

A fully connected layer of a neural network is defined by a $N \times M$ matrix $W$ of column vectors $w_i$, bias $b$, and the equation $y = W^T x + b$; $N$ is the number of classes, and $M$ is the dimension of the ``representation space'' containing $x$. In a classifier, such as we use, the elements $y_i$ are subject to the softmax function $z_i = \exp( y_i ) / \sum_j \exp( y_j )$ that exponentially emphasises any dominant direction. The loss function used is cross-entropy. Once trained, the resulting vector $z = [z_1, ..., z_N]$ can then be used to classify the point $x$ by selecting the dominant direction, the class index being given by: $k^* = \argmax_k z_k$.

We fix the final layer to random weights, so that the $w_i$ are fixed basis vectors. Recalling that we define class to be a direction, this is sufficient to for our purpose. Our training is of network weights up to the last layer, which forces input images in a given class to map to some point $x$ such that its class direction $w_i$ is dominant. Different depiction instances are free to spread along that direction by any amount, provided that it remains the dominant direction. This allows for a wide variation in style -- as spread along a class basis vector -- classes are separated by the angle between basis vectors.





In general, randomly chosen weights for $W$ will not produce $w_i$ that are orthogonal, and neither will they be the same length. We can employ singular value decomposition $W = U \Sigma V^T$.  keeping the first $N$ vectors of the $M \times M$ unitary matrix $V$ as the random projection. In practise, we find high dimensions $M$ give basis vectors $w_i$ that are near-orthogonal and of near-equal length.

\begin{algorithm}
\begin{algorithmic}[1]
\Require \text{$N_{iter}$: Number of training iterations}
\Require \text{$\alpha, \beta, \gamma$: Learning rate hyperparameters}
\For{$t$ in $1:N_{iter}$}
\State Split $p$ train domains into meta-train $a$ and meta-tests $b$
\State Sample \textit{meta-train} set $\{(x_j^{(a)}, y_j^{(a)})\sim D_a\}_{j=1}^{n_b}$
\State Perform supervised classification updates with meta-train
\State Meta-train gradients $\nabla_\Theta = \mathcal{F}'_\Theta(\mathcal{\bar{S}};\Theta)$
\State Meta-updated parameters $\Theta'=\Theta - \alpha\nabla_\Theta$
\State Sample \textit{meta-test} set $\{(x_j^{(b)}, y_j^{(b)})\sim D_b\}_{j=1}^{n_b}$
\State Meta-test loss $\mathcal{G(\hat{S}}; \Theta')$
\State Meta-optimisation $\Theta = \Theta - \gamma\frac{\partial(\mathcal{F(\bar{S}};\Theta) + \beta\mathcal{G(\hat{S}};\Theta - \alpha\nabla_\Theta)}{\partial\Theta}$
\EndFor
\end{algorithmic}
\caption{MLDG Training Algorithm \cite{mldg}}
\label{alg:mldg}
\end{algorithm}

\begin{algorithm}
\begin{algorithmic}[1]
\Require \text{$N_{iter}$: Number of training iterations}
\Require \text{$\alpha_1, \alpha_2$: Learning rate hyperparameters}
\For{$t$ in $1:N_{iter}$}
\For{$i$ in $1:p$}
\State \text{Sample $n_b$ labelled images $\{(x_j^{(i)}, y_j^{(i)})\sim D_i\}_{j=1}^{n_b}$}
\State \text{Perform supervised classification updates:}
\State \text{$\psi^{(t)} \leftarrow \psi^{(t-1)} - \alpha_1\nabla_\psi L^{(i)}(\psi^{(i)}, \theta_i^{(t-1)})$}
\State \text{$\theta_i^{(t)} \leftarrow \theta_i^{(t-1)} - \alpha_1\nabla_{\theta_i} L^{(i)}(\psi^{(i)}, \theta_i^{(t-1)})$}
\EndFor
\State Choose $a, b\in \{1, 2, ... p\}$ randomly such that $a \neq b$
\State $\beta^1 \leftarrow\ \theta_a^{(t)}$
\For{$i=1:l$}
\State Sample \textit{meta-train} set $\{(x_j^{(a)}, y_j^{(a)})\sim D_a\}_{j=1}^{n_b}$
\State $\beta^i = \beta^{i-1} - \alpha_2\nabla_{\beta^{i-1}}[L^{(a)}(\psi^{(t)} \beta^{i-1}) + R_\theta(\beta^{i-1})]$
\EndFor
\State $\hat{\theta}_a^{(t)} = \beta_l$
\State Sample \textit{meta-test} set $\{(x_j^{(b)}, y_j^{(b)})\sim D_b\}_{j=1}^{n_b}$
\State Perform meta-update for regulariser $\phi^{(t)} = \phi^{(t-1)} - \alpha_2\nabla_\phi L^{(b)}(\psi^{(t)}, \hat{\theta}_a^{(t)})|_{\phi=\phi^{(t)}}$
\EndFor
\end{algorithmic}
\caption{MetaReg Training Algorithm \cite{metareg}}
\label{alg:metareg}
\end{algorithm}

\begin{algorithm}
\begin{algorithmic}[1]
\Require \text{$N_{iter}$: Number of training iterations}
\Require \text{$\alpha$: Learning rate hyperparameters}
\For{$t$ in $1:N_{iter}$}
\State Sample aggregation of training domains $\{(x_j, y_j)\sim D\}_{j=1}^{n_b}$
\State Perform supervised classification updates with sampled batch
\State Gradients $\nabla_\Theta = \mathcal{F}'_\Theta(\mathcal{\bar{S}};\Theta)$
\State Updated parameters $\Theta'=\Theta - \alpha\nabla_\Theta$
\EndFor
\end{algorithmic}
\caption{Our Training Algorithm (Standard SGD)}
\label{alg:us}
\end{algorithm}

\section{Results}
\label{sec:expts}

We compare our method to MLDG~\cite{mldg} and MetaReg~\cite{metareg} using the PACS benchmark \cite{dbadg}. To provide comparison with other domain generalisation approaches with include results; from Domain Separation Network (DSN) \cite{dsn}; from Domain Multi-Task Auto Encoders (D-MTAE) \cite{dmtae}; and from (DBA-DG) ~\cite{dbadg}, all of which have been used on PACS (see~\cite{dbadg} and note DSN is a domain adaptation method Li \etal \cite{dbadg} modified for generalisation). We conform to the standard approach for using PACS, described in Subsection~\ref{subsec:PACS}, of using three source domains for training and one held-out domain for testing with our ``fixed random weight'' classifier.

To make comparisons as direct and fair as possible, we follow recent work on PACS (including DBA-DG \cite{dbadg}, MLDG \cite{mldg} and MetaReg \cite{metareg}) in using AlexNet \cite{alexnet} as the base architecture. In the literature, the baseline setup for the PACS benchmark consists of training all layers of AlexNet on the aggregation of training domains \cite{dbadg,mldg} (referred to as ``Full AlexNet'' here). We recreate the training hyperparameters of the literature \cite{mldg}: batch size $64$, learning rate $5e-4$ with exponential decay $0.96$ every $15k$ steps. As in AlexNet's \cite{alexnet} original training we include momentum $0.9$ and weight decay $5e-5$. We train our baseline models with ImageNet pretrained weights for initialisation and subsequent models using our baseline weights for initialisation. PACS training domains are split 9:1 into training and validation sets, Li \etal~\cite{dbadg, mldg} select the best performing model on the validation set after 45k iterations and deploy this onto the test domain.

We report the performance of our model with random weights for the final fully connected layer, \textit{fc8}, of Alexnet. Our models are implemented in PyTorch, weights sampled from a random uniform distribution $\ \mathcal{U}(a, b)$ where $a, b$ are lower and upper bounds. We use the default PyTorch initialisation where $a=-b$, $b=1/\sqrt{N_n}$ and $N_n$ is the number of neurons in the input layer ($N_n=4096$ in AlexNet).

\begin{table*}[h]
\begin{center}
\begin{tabular}{c|c|c|c|c|c}
 & Art Painting & Cartoon & Photo & Sketch & Average\\
\hline
Full AlexNet (Ours) & 61.18 & 65.70 & 88.14 & 55.95 & 67.74 \\
D-MTAE \cite{dmtae} & 60.27 & 58.65 & \textbf{91.12} & 47.68 & 64.48 \\
DSN \cite{dsn} & 61.13 & 66.54 & 83.25 & 58.58 & 67.37 \\
DBA-DG \cite{dbadg} & 62.86 & 66.97 & 89.50 & 57.51 & 69.21 \\
MLDG \cite{mldg} & 66.23 & 66.88 & 88.00 & 58.96 & 70.01 \\
MetaReg \cite{metareg} & \textbf{69.82} & \textbf{70.35} & 91.07 & 59.26 & \textbf{72.62} \\
Ours & 60.25 & 68.38 & 88.27 & \textbf{63.01} & 70.00 \\
Ours Orthogonal & 60.81 & 67.46 & 87.96 & 61.96 & 69.55\\
\hline
\end{tabular}
\caption{Cross-depiction recognition accuracy (in \%) on the PACS dataset. }
\label{PACS_1}
\end{center}
\end{table*}

Table \ref{PACS_1} reports our results. All three methods (Ours, MLDG \cite{mldg} and MetaReg \cite{metareg}) produce a similar $2$-$3\%$ improvement over their respective ``Full AlexNet'' baselines. Our classifier compares well with each of the comparator algorithms, despite ours being the least complex algorithm. Forcing orthogonal class basis vectors makes little difference to our results. MetaReg~\cite{metareg} is by far the most complex algorithm, and although it achieves maximum performance on all styles (other than ``sketches, where we register the best performance) it still exhibits a significant fall in performance for all ``arty'' styles compared to photographs. Such a fall is witnessed not just for MetaReg \cite{metareg} but for all algorithms in our table, and our results are consistent with those reported elsewhere -- the “cross depiction” problem can be regarded as the challenge to maintain classification performance across depictions.

\section{Discussion and Conclusion}
\label{sec:conc}

Experimental results show that a simple classifier with fixed random weights performs comparably well to far more sophisticated algorithms. It is of course true that those more sophisticated algorithms were not designed for the cross-depiction {\em per se} (but neither was ours) and it is also true that our tests were limited to a single dataset comprising discrete and very distinct styles. Even so, the results obtained here echo the results obtained elsewhere: computer vision algorithms tend to perform well on photographic imagery but not on artistic imagery. More exactly, networks trained on one depiction ``domain'' do not generalise well to other ``domains''. Our experiments show this is true even in the case of meta-learning. In general, a good learning measure is valuable; however, it does not necessarily provide a meaningful improvement in the cross-depiction problem. 

\begin{figure}[h]{}
\begin{center}
\includegraphics[height=2.5cm]{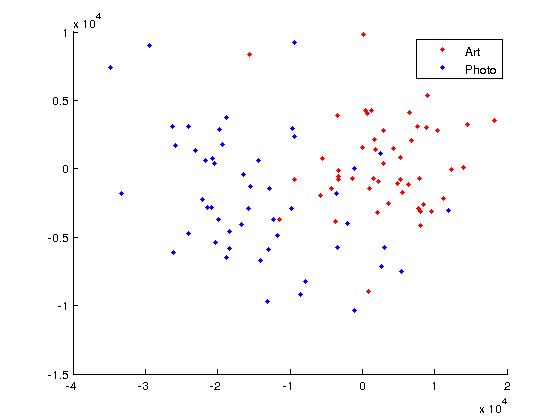}
\includegraphics[height=2.5cm]{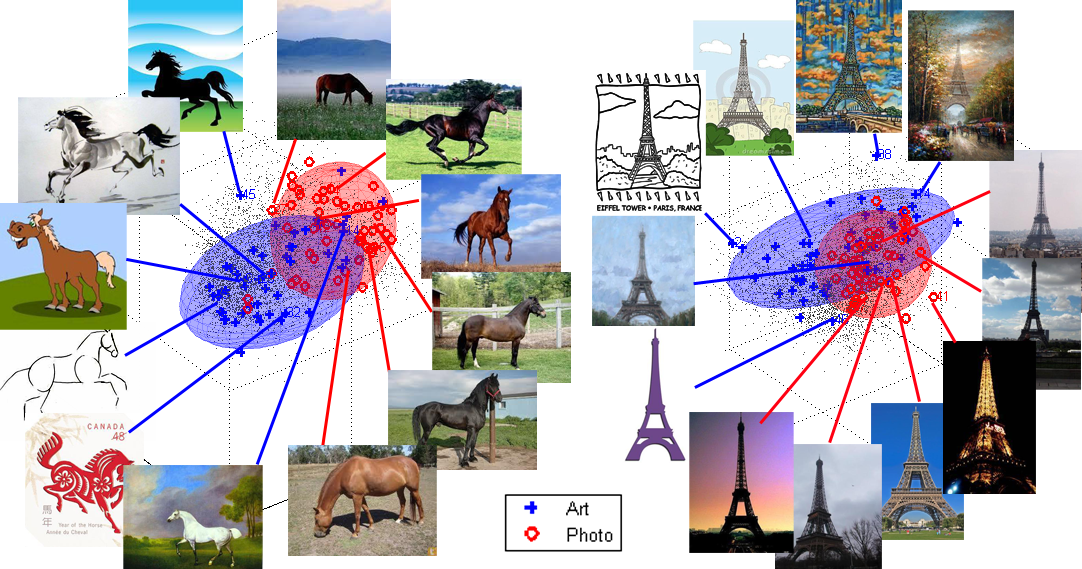}
\includegraphics[height=2.5cm]{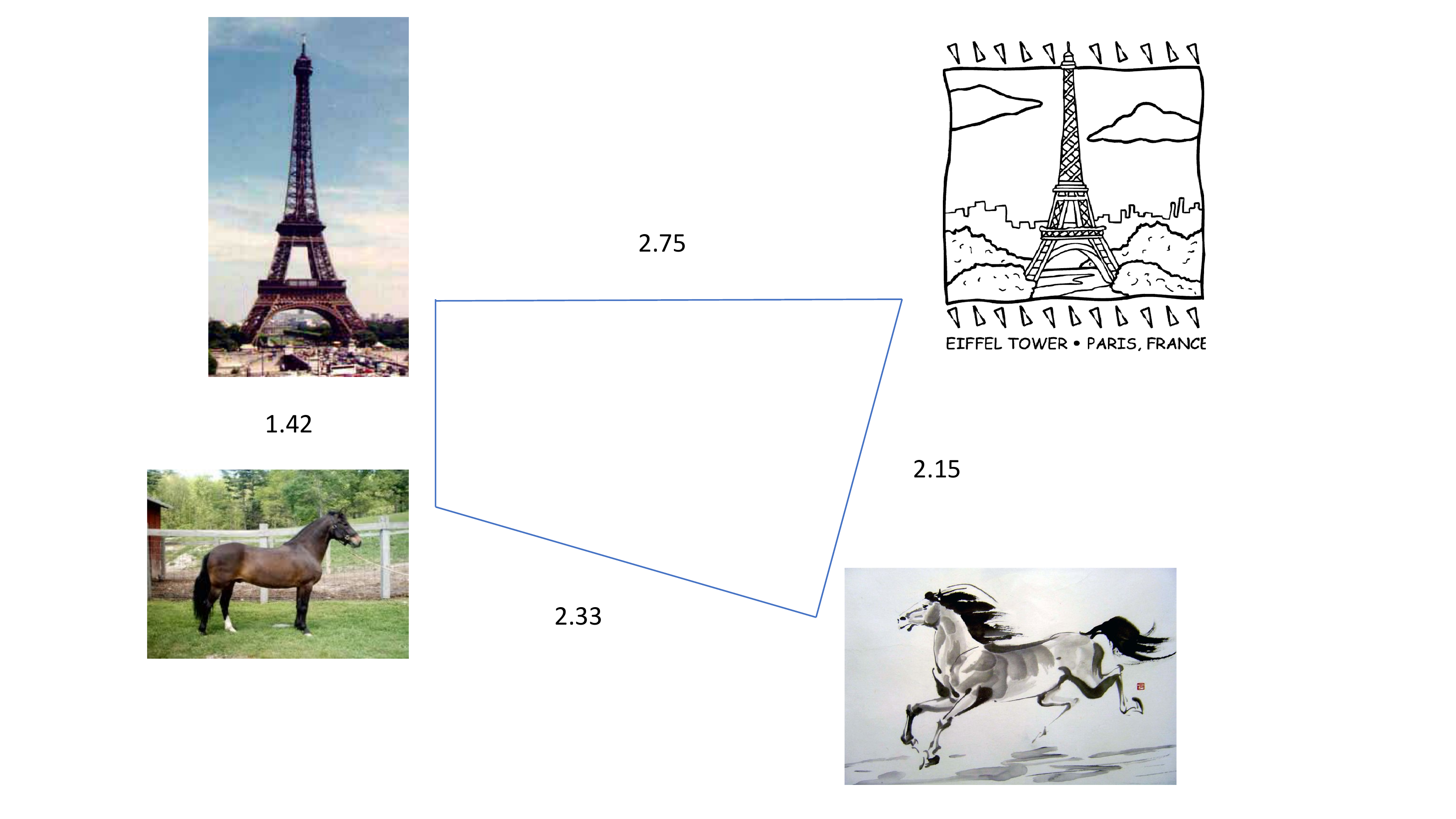}
\end{center}
\caption{Visual class centres for artwork (red) and photographs (blue)  -- pictures in the same class but different depictions tend to be further apart than pictures in the same depiction but different class (taken from \protect{\cite{Hall2015}}).}
\label{fig:faceNspace}
\end{figure}

\begin{figure}[h]{}
\begin{center}
\includegraphics[height=2.5cm]{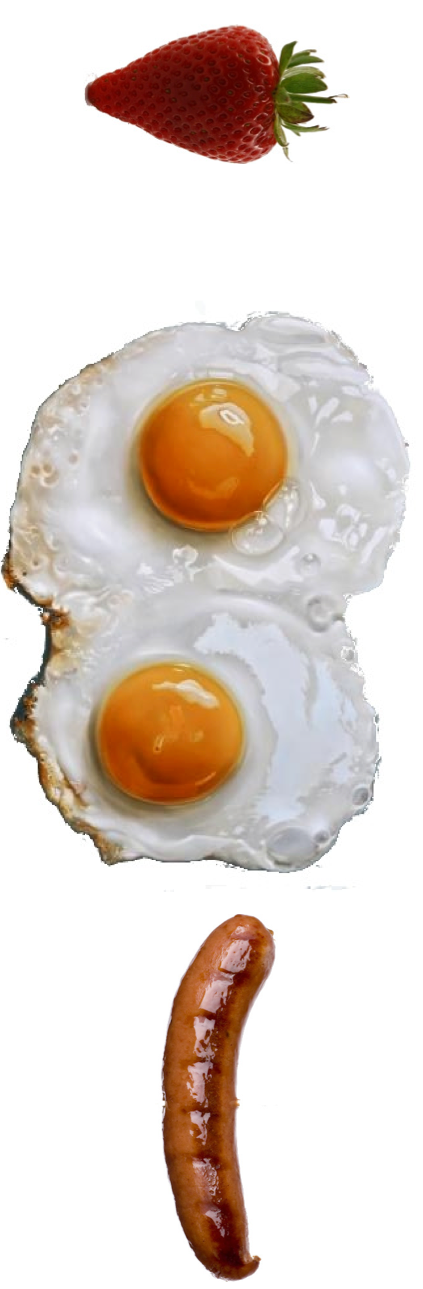}
\qquad
\includegraphics[height=2.5cm]{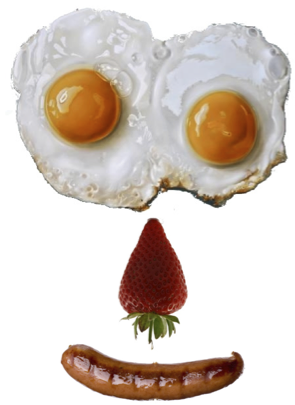}
\end{center}
\caption{Three foods, photographed make a face when properly arranged in space.}
\label{fig:breakfast}
\end{figure}

Recognition premised primarily on texture will fail to represent class models in a manner that generalises across depictions. Using an example from Hall {\em et al}~\cite{Hall2015}, Figure ~\ref{fig:faceNspace} shows the the centre of visual classes projected onto the largest two eigenvectors of ``image space'' (for $M \times N$ RGB images this is an $M \times N \times 3$ dimensional space  in which each picture is a point specified by RGB-values at each pixel). In fact each class has two centres, one photographic (blue) the other artistic (red): the photographs and artworks tend to separate out. Looking at specific classes (horse, Eiffel tower) illustrates the spread of depictions over classes, while individual images show that objects in the same class but different depictions are further apart than different objects in the same depiction. Equally, Figure \ref{fig:breakfast} illustrates how the spatial arrangement of parts impacts significantly on the ability of humans to recognise. 

We conclude that the failure of networks, and indeed conventional algorithms, to generalise across depictions needs an explanation. We contend that the problem of cross-depiction is not one of measure alone: no matter how sophisticated a measure is used, the problem of recognition is primarily one of representation. The only recognition algorithm we know of that does not exhibit a fall in performance across depictions is that due to Wu {\em et al}~\cite{Wu2014}, which used a fully-connected weighted graph model as its base class model. Furthermore, recent work has shown that networks over-rely on texture for recognition and highlighted the importance of shape~\cite{texturebias}. That is, structure and shape are fundamental to object class recognition. The cross-depiction problem remains challenging exactly because it pushes at the basic assumptions commonly made.

\FloatBarrier
\bibliographystyle{ieee}
\bibliography{egbib}
\end{document}